\documentclass{article}
\usepackage{spconf,amsmath,graphicx,hyperref}
\usepackage{float}
\usepackage{flafter}
\usepackage{amssymb}
\usepackage{xcolor}
\usepackage{booktabs}
\hypersetup{hidelinks}
\newcommand{\xs}[1]{\textcolor{black}{#1}}

\title{Representation Disentanglement for Fair Chest X-Ray Diagnosis}
\name{
Yujie Sun$^{1}$,
Ruizhe Li$^{2}$,
Xiaowu Sun$^{1}$
}

\address{\fontsize{11.5}{13}\selectfont
$^{1}$ Academy of Artificial Intelligence and Advanced Technology,
Xi'an Jiaotong-Liverpool University, China \\
\fontsize{11.5}{13}\selectfont
$^{2}$ Nottingham Biomedical Research Centre, School of Medicine, University of Nottingham, UK
}
\begin{document}
%
\maketitle
%



\begin{abstract}
Deep learning has advanced chest X-ray (CXR) diagnosis, yet demographic biases in learned representations may contribute to performance disparities across intersectional groups. We propose a single-encoder framework combining dual-level decorrelation with prototype-guided cross-group contrastive learning to reduce demographic dependence while accounting for within-class variation. We further propose Demographic Representation Alignment Reduction (DRAR), a new metric that quantifies the reduction in demographic structure within disease representations. The framework is evaluated on four classification tasks using 34,809 CheXpert test images across eight intersectional groups, defined by age, sex and ethnicity. Compared with empirical risk minimization (ERM), our method reduces the mean equalized-odds gap from 15.41\% to 10.86\% and the AUC gap from 5.95\% to 5.01\%. Our method achieves a DRAR of 59.04\% relative to ERM, with only a slight decrease in mean AUC. These results demonstrate that representation disentanglement can reduce demographic bias and improve intersectional fairness.
Code is available at \url{https://github.com/06Yujie/Fair-Medical-Imaging}.
\end{abstract}


\section{Introduction}
\label{sec:introduction}
Advances in chest X-ray (CXR) representation learning have improved disease recognition~\cite{yao2025evax}. However, demographic bias remains a critical concern in CXR diagnosis~\cite{xu2024fairnessreview,glocker2023risk}. Models can exploit demographic and acquisition-related cues that correlate with disease labels. Reliance on these cues can encourage shortcut learning and impair generalization under distribution shifts~\cite{lotter2024acquisition,yang2024limits}. Such shortcuts can contribute to uneven diagnostic performance across demographic groups.

Several strategies have been developed to reduce demographic bias in disease diagnosis, including data-level augmentation~\cite{drexlin2025medi} and bias-aware optimization~\cite{luo2025adaabc}. More recently, representation-based methods have learned separate disease and demographic representations and disentangled them during training~\cite{zhao2025dfca,gao2026fairread,yi2025adfairclip}. However, such methods often rely on separate encoders, complicating joint learning. In addition, cross-group alignment based on disease labels may overlook within-class variation, as patients sharing the same label can exhibit heterogeneous imaging patterns~\cite{zhao2025dfca}. Uniform alignment may therefore suppress diagnostically relevant variation and affect performance differently across demographic groups~\cite{gao2026fairread}. These limitations motivate a unified framework that jointly learns and disentangles disease and demographic representations while accounting for within-class patterns during cross-group alignment.

Moreover, evaluating fairness from model predictions provides limited insight into the learned representation space. Subgroup metrics, such as equalized-odds and area under the receiver operating characteristic curve (AUC) disparities, quantify performance gaps across demographic groups~\cite{hardt2016equality}, while demographic probing measures the recoverability of sensitive attributes~\cite{yang2024limits}. However, neither directly quantifies the correspondence between disease representation geometry and demographic structure. This motivates a representation-level metric that captures such correspondence, complementing performance-based fairness evaluation.

To address these challenges, we propose a unified framework that disentangles disease and demographic representations using a single encoder. We further account for within-class variation during cross-group alignment and introduce representation-level evaluation of demographic bias. We evaluate the framework across intersectional groups defined by age, sex, and ethnicity. Our main contributions are:

\par
\begingroup
\setlength{\parskip}{0pt}

\noindent\hspace*{0em}\textbullet\enspace
We develop a single-encoder framework with feature- and relation-level decorrelation to disentangle disease and demographic representations.
\par
\noindent\hspace*{0em}\textbullet\enspace
We propose prototype-guided cross-group contrastive learning to account for within-class variation during alignment.
\par
\noindent\hspace*{0em}\textbullet\enspace
We introduce a representation-level fairness metric that quantifies demographic structure retained in disease representations.
\par
\endgroup

\vspace{-3mm}
\section{Methodology}
\label{sec:method}
\begin{figure*}[h]
    \centering
    \includegraphics[width=0.9\textwidth]{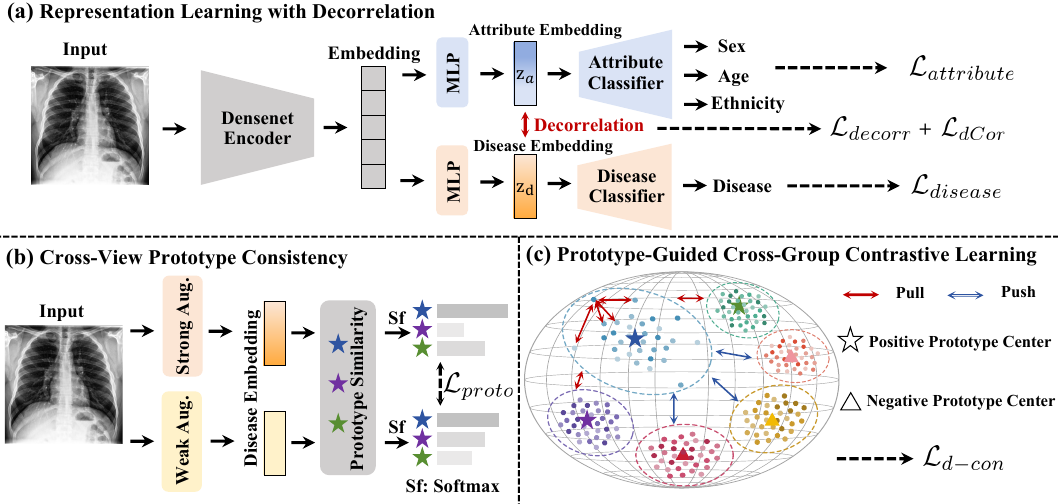}
    \vspace{-0.3cm}
    \caption{Overview of the proposed framework. A shared encoder learns disease and attribute spaces through prototype-guided cross-group alignment and dual-level decorrelation.}
    \label{fig:pipeline}
\vspace{-0.5cm}
\end{figure*}

The framework has three main components: disease and demographic representation learning with an encoder and heads, feature- and relation-level decorrelation for representation disentanglement, and prototype-guided cross-group alignment that accounts for within-class variation.
\vspace{-4mm}
\subsection{Disease and Demographic Representation Learning}

Given an image, a single DenseNet encoder~\cite{huang2017densely} extracts a common representation, which is mapped by separate projection heads to $\ell_2$-normalized disease and demographic representations $\mathbf{z}_{d,i}$ and $\mathbf{z}_{a,i}$, respectively, as shown in Fig.~\ref{fig:pipeline}.a.

The two representation spaces are learned through task-specific supervision. The disease representation is supervised by binary cross-entropy $\mathcal{L}_{\mathrm{disease}}$, while the demographic representation is supervised by the averaged cross-entropy loss $\mathcal{L}_{\mathrm{attribute}}$ for age, sex, and ethnicity. This supervision encourages the two heads to encode disease and demographic information in their respective representation spaces. During inference, only the encoder, disease projection head, and disease classifier are retained.
\vspace{-3mm}
\subsection{Feature- and Relation-Level Decorrelation}

Separate projection heads do not ensure disentanglement of disease and demographic information. \xs{As shown in Figure~\ref{fig:pipeline}.a,} we therefore reduce their dependence at two complementary levels: feature-level decorrelation suppresses associations between feature dimensions, while relation-level decorrelation reduces dependence between pairwise distance patterns in the two spaces.

\textbf{Feature-level decorrelation.}
To suppress feature-level dependence without penalizing between-class disease differences, we center both representations within each disease class and minimize their class-conditional centered cross-product:
\vspace{-2mm}
\begin{equation}
\mathcal{L}_{\mathrm{decorr}}=
\frac{1}{|\mathcal{C}|h^2}
\sum_{c\in\mathcal{C}}
\left\|
\widetilde Z_d^{(c)}
\operatorname{sg}(\widetilde Z_a^{(c)})^\top
\right\|_F^2 ,
\label{eq:decorr}
\vspace{-2mm}
\end{equation}
where $\mathcal{C}\subseteq\{0,1\}$ is the set of disease classes represented by at least two samples in the mini-batch, and $\widetilde{\mathbf{Z}}_d^{(c)}, \widetilde{\mathbf{Z}}_a^{(c)}\in\mathbb{R}^{h\times n_c}$ are the corresponding class-centered disease and demographic representation matrices, with samples arranged as columns. Here, $h$ is the representation dimension, $\operatorname{sg}(\cdot)$ denotes stop-gradient, and $\|\cdot\|_F$ is the Frobenius norm.

\textbf{Relation-level decorrelation.}
Feature-level decorrelation does not necessarily remove demographic structure from the geometry of the disease representation. We therefore characterize sample relationships using pairwise Euclidean distance matrices $D_d,D_a\in\mathbb{R}^{n\times n}$, with entries $D_d^{ij}=\|\mathbf{z}_{d,i}-\mathbf{z}_{d,j}\|_2$ and $D_a^{ij}=\|\mathbf{z}_{a,i}-\mathbf{z}_{a,j}\|_2$ for a mini-batch of $n$ samples, where $i,j\in\{1,\ldots,n\}$. Each matrix is double-centered by subtracting its row
and column means and adding its grand mean, yielding $A_d$ and $A_a$. We minimize
\vspace{-2mm}
\begin{equation}
\mathcal{L}_{\mathrm{dCor}}
=
\frac{
\max\!\left(
0,
\left\langle A_d,\operatorname{sg}(A_a)\right\rangle_F
\right)
}{
\sqrt{
\left\langle A_d,A_d\right\rangle_F
\left\langle
\operatorname{sg}(A_a),\operatorname{sg}(A_a)
\right\rangle_F
+\epsilon
}
},
\label{eq:dcor}
\end{equation}
where $\langle\cdot,\cdot\rangle_F$ is the Frobenius inner product, $\max(0,\cdot)$ avoids numerical negatives, and $\epsilon$ ensures stability. Minimizing $\mathcal{L}_{\mathrm{dCor}}$ reduces dependence between pairwise distance patterns in the two spaces.
\vspace{-3mm}
\subsection{Prototype-Guided Variation-Aware Cross-Group Alignment}
Beyond decorrelation, disease representations should be consistent across demographic groups while preserving meaningful within-class variation. Uniformly aligning same-label samples may suppress such variation. \xs{We therefore use class-specific prototypes~\cite{reiss2023decoupled} to characterize within-class patterns and guide cross-group alignment between samples with similar disease-related characteristics.}

To model heterogeneous within-class patterns, we use $K=3$ learnable unit-norm prototypes $\{\mathbf{u}_{c,k}\}_{k=1}^{K}$ for each class $c\in\{0,1\}$ in the disease space. For sample $i$, a softmax with temperature $\tau_p$ converts its cosine similarities to the prototypes of class $y_i$ into an assignment $\mathbf{p}_i=(p_{i1},\ldots,p_{iK})^\top$, representing its affinity to different within-class patterns. \xs{As shown in Fig.~\ref{fig:pipeline}.b,} we enforce consistency of these assignments across augmented views and combine prototype affinity with demographic differences to weight pairs for contrastive alignment.

\textbf{Cross-view prototype consistency.}
To obtain reliable within-class pattern assignments, we construct weak and strong views $x_i^w$ and $x_i^s$ using cropping, resizing, flipping, and affine transformations, with additional brightness and contrast jitter and Gaussian blur for the strong view. Both views are processed by the same encoder and disease projection head to obtain prototype assignments $\mathbf{p}_i^w$ and $\mathbf{p}_i^s$. Using the weak-view assignment as the target, we minimize
\vspace{-3mm}
\begin{equation}
\mathcal{L}_{\mathrm{proto}}
=
\frac{1}{n}\sum_{i=1}^{n}
D_{\mathrm{KL}}
\left(
\operatorname{sg}(\mathbf{p}_i^{w})
\,\middle\|\,
\mathbf{p}_i^{s}
\right),
\label{eq:prototype-consistency}
\vspace{-3mm}
\end{equation}
where $n$ is the mini-batch size and $D_{\mathrm{KL}}$ denotes the Kullback--Leibler divergence.

\textbf{Prototype-guided pair weighting.}
\xs{As illustrated in Fig.~\ref{fig:pipeline}.c,} samples sharing the same disease label form positive pairs, while those with different labels form negative pairs. For pair $(i,j)$, we define prototype affinity $r_{ij}$ and demographic difference $\delta_{ij}$ as
\vspace{-3mm}
{\small
\begin{equation}
r_{ij}=1-\rho+\rho\,\operatorname{sg}(\mathbf{p}_i^s)^\top\operatorname{sg}(\mathbf{p}_j^s),\quad
\delta_{ij}=\frac{1}{3}\sum_{t=1}^{3}\mathbf{1}[a_{it}\ne a_{jt}],
\label{eq:pair-affinity}
\end{equation}
}
where \xs{$t\in \{1,2,3\}$ indexes the three demographic attributes, and} $a_{it}$ denotes the $t$th demographic attribute of sample $i$, $\mathbf{1}[\cdot]$ is the indicator function, and $\rho$ controls the contribution of prototype affinity. Their pair weights are
\begin{equation}
\begin{aligned}
w^{+}_{ij}&=\mathbf{1}[y_i=y_j](r_{ij}+\epsilon)^{\gamma}(1+\alpha\delta_{ij}),\\
w^{-}_{ij}&=\mathbf{1}[y_i\ne y_j]\exp(-\delta_{ij}/\sigma),
\end{aligned}
\label{eq:pairweights}
\end{equation}
where $\alpha$, $\gamma$, and $\sigma$ control cross-group emphasis, prototype selectivity, and negative-pair decay, respectively, and $\epsilon$ ensures numerical stability. Thus, positive pairs with similar within-class patterns and greater demographic differences receive larger weights, whereas negative pairs with similar demographics are emphasized.

\textbf{Prototype-guided contrastive learning.}
Using these pair weights, we perform supervised contrastive learning in the disease representation space. Let $q_{ij}=(\bar{\mathbf{z}}_{d,i}^{s})^\top \bar{\mathbf{z}}_{d,j}^{s}/\tau$ denote the temperature-scaled cosine similarity, where $\bar{\mathbf{z}}_{d,i}^{s}
=\mathbf{z}_{d,i}^{s}/\lVert\mathbf{z}_{d,i}^{s}\rVert_2$
and $\tau$ is the contrastive temperature. For each anchor $i$, let $\mathcal{P}(i)=\{j\ne i:y_j=y_i\}$ denote its positive set. We normalize the positive-pair weights as $\pi_{ij}=w^+_{ij}/\sum_{\ell\in\mathcal{P}(i)}w^+_{i\ell}$ for $j\in\mathcal{P}(i)$, and define the weighted normalization term $b_i=\sum_{j'\ne i}(w^+_{ij'}+w^-_{ij'})\exp(q_{ij'})$. The contrastive loss is
\begin{equation}
\mathcal{L}_{d\text{-}con}
=
-\frac{1}{|\mathcal{I}|}
\sum_{i\in\mathcal{I}}
\sum_{j\in\mathcal{P}(i)}
\pi_{ij}
\log
\frac{w^+_{ij}\exp(q_{ij})}{b_i},
\label{eq:disease-contrastive}
\end{equation}
where $\mathcal{I}=\{i:\mathcal{P}(i)\ne\emptyset\}$ contains anchors with at least one positive pair. This objective promotes disease-consistent representations across demographic groups while accounting for within-class variation.

The complete training objective combines disease and demographic supervision, dual-level decorrelation, prototype consistency, and contrastive alignment:
\vspace{-2mm}
\begin{equation}
\begin{aligned}
\mathcal{L}={}&\mathcal{L}_{\mathrm{disease}}
+\lambda_a\mathcal{L}_{\mathrm{attribute}}
+\mathcal{L}_{d\text{-}con}
+\mathcal{L}_{\mathrm{proto}}\\
&+\lambda_{\mathrm{decorr}}\mathcal{L}_{\mathrm{decorr}}
+\lambda_{\mathrm{dCor}}\mathcal{L}_{\mathrm{dCor}},
\end{aligned}
\vspace{-2mm}
\label{eq:total}
\end{equation}
where $\lambda_a$, $\lambda_{\mathrm{decorr}}$, and $\lambda_{\mathrm{dCor}}$ weight attribute supervision, feature-level decorrelation, and relation-level decorrelation.
\vspace{-3mm}
\subsection{Representation-Level Fairness Metric}

Existing fairness evaluation does not directly characterize demographic structure in disease representations. We therefore propose \textbf{D}emographic \textbf{R}epresentation \textbf{A}lignment \textbf{R}eduction (DRAR), which measures the structural alignment between disease representations and a demographic reference space and quantifies its reduction relative to ERM.

We train a reference model jointly supervised by age, sex, and ethnicity to obtain a demographic reference representation $Z_a^{\mathrm{ref}}$. For each evaluated model, we measure the structural alignment between its disease representation $Z_d$ and the demographic reference using centered kernel alignment (CKA)~\cite{kornblith2019similarity}, denoted by $S=\mathrm{CKA}(Z_d,Z_a^{\mathrm{ref}})$. DRAR is then defined as
\vspace{-4mm}
\begin{equation}
\mathrm{DRAR}=\frac{S_{\mathrm{ERM}}-S}{S_{\mathrm{ERM}}}.
\label{eq:drar}
\end{equation}
ERM has $\mathrm{DRAR}=0$, while higher values indicate a greater reduction in demographic alignment. DRAR complements outcome-level fairness metrics and demographic probing.
\vspace{-4mm}
\section{Experiments and results}
\label{sec:experiments}

\subsection{Dataset and Demographic Groups}
\label{sec:dataset}
We use the CheXpert dataset~\cite{irvin2019chexpert}, which comprises 224,316 chest radiographs from 65,240 patients. Four binary classification tasks are considered: \textit{No Finding}, \textit{Cardiomegaly}, \textit{Pleural Effusion}, and \textit{Pneumothorax}. The test set consists of 34,809 images and is divided into eight intersectional groups based on sex (Female/Male), ethnicity (White/Non-White), and age ($<60$/$\geq60$ years).
\vspace{-3mm}
\subsection{Experimental Setup}
\label{sec:experiment}
We compare our method with ERM, ReSample~\cite{idrissi2022resample}, GroupDRO~\cite{sagawa2020groupdro}, DANN~\cite{ganin2016dann}, FairAdaBN~\cite{xu2023fairadabn}, and FairREAD~\cite{gao2026fairread}. For our method, we evaluated $K\in\{1,3,5\}$ and selected $K=3$ based on the validation set AUC, we further set $h=128$, $\tau=0.1$, $\tau_p=0.2$, $\rho=\alpha=\gamma=1$, $\sigma=0.5$, $\epsilon=10^{-12}$, $\lambda_a=0.25$, $\lambda_{\mathrm{decorr}}=500$, and $\lambda_{\mathrm{dCor}}=1$.

\vspace{-4mm}
\subsection{Evaluation Metrics}
We evaluate fairness and diagnostic utility at both prediction and representation levels:

\textbf{Prediction level:} Diagnostic utility is evaluated using balanced accuracy (BACC), area under the receiver operating characteristic curve (AUC), and macro F1 score. Intersectional fairness across the eight demographic groups is measured by AUC disparity ($\Delta_{\mathrm{AUC}}$) and equalized-odds disparity ($\Delta_{\mathrm{EO}}$). Their trade-off is further assessed using $\mathrm{FATE}_{\mathrm{AUC}}$ and $\mathrm{FATE}_{\mathrm{EO}}$~\cite{xu2023fairadabn}, with BACC as the utility measure.

\textbf{Representation level:} Besides our proposed DRAR metric, age, sex, and ethnicity are predicted from frozen disease embeddings using auxiliary classifiers, with lower accuracy indicating less demographic information.


\vspace{-3.5mm}
\subsection{Results}
\vspace{-3mm}

We compare the methods at prediction and representation levels. We assess diagnostic utility and intersectional fairness, demographic information retained in disease representations, and cross-group representation patterns. Ablation studies further examine the contributions of the proposed components.

\textbf{Diagnostic Utility and Intersectional Fairness.}
Table~\ref{tab:main_results} reports the four-task mean results on the test cohort. Our method achieves the lowest disparities, reducing $\Delta_{\mathrm{EO}}$ from 15.41\% to 10.86\% and $\Delta_{\mathrm{AUC}}$ from 5.95\% to 5.01\% relative to ERM. Meanwhile, AUC decreases only from 83.80\% to 83.36\% and BACC from 76.44\% to 75.47\%. These results show that demographic disparities are substantially reduced while diagnostic utility is largely preserved, consistent with our objective of improving fairness rather than aggregate classification performance. Accordingly, our method achieves the highest $\mathrm{FATE}_{\mathrm{EO}}$ and $\mathrm{FATE}_{\mathrm{AUC}}$, indicating the best utility--fairness trade-off among the compared methods.

\begin{figure}[!t]
\centering
\includegraphics[width=\columnwidth]{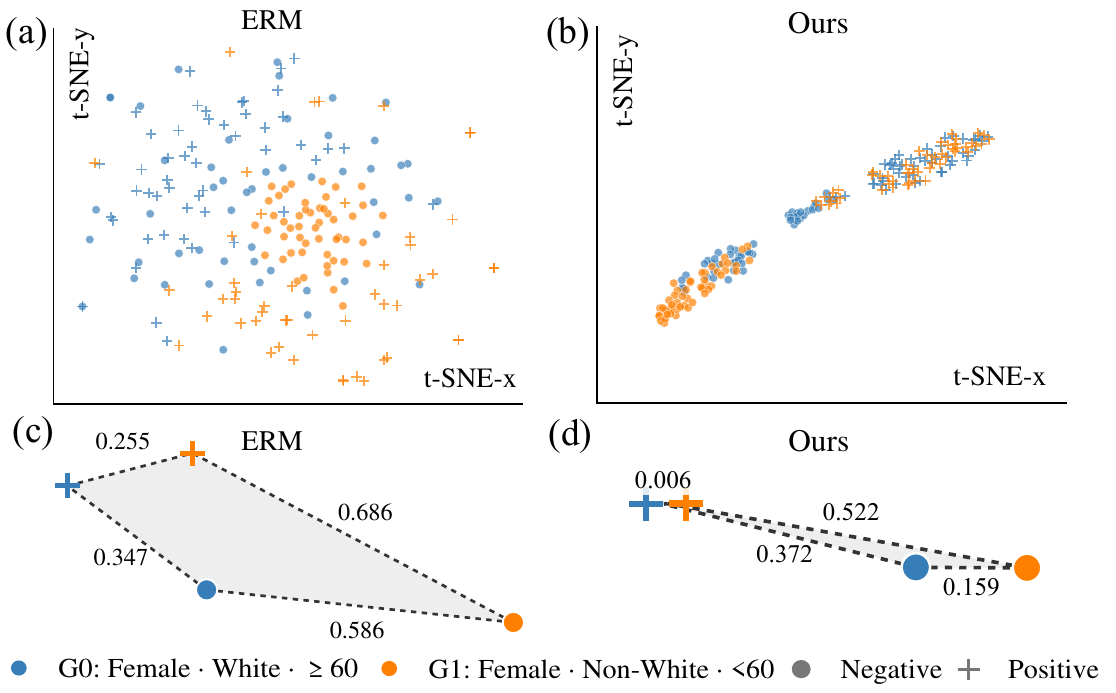}
\vspace{-8mm}
\caption{Pleural Effusion feature analysis. (a,b) t-SNE plots for ERM and our method. (c,d) Mean Euclidean distances between group--label representations in original feature space.}
\vspace{-4mm}
\label{fig:feature_analysis}
\end{figure}

\begin{table}[ht]
\vspace{-0.5cm}
\centering
\caption{Four-task mean diagnostic and intersectional fairness results (\%). $F$ denotes the corresponding FATE score.}
\label{tab:main_results}
\renewcommand{\arraystretch}{1.05}
\setlength{\tabcolsep}{3.5pt}
\resizebox{\columnwidth}{!}{%
\begin{tabular}{lccccccc}
\noalign{\hrule height 0.8pt}
Method & BACC $\uparrow$ & AUC $\uparrow$ & F1 $\uparrow$ & $\Delta_{\rm EO}\downarrow$ & $\Delta_{\rm AUC}\downarrow$ & $F_{\rm EO}\uparrow$ & $F_{\rm AUC}\uparrow$ \\
\hline
ERM       & \textbf{76.44} & \textbf{83.80} & 66.22          & 15.41          & 5.95          & 0.00           & 0.00 \\
GroupDRO  & 74.99          & 82.07          & 65.22          & 12.43          & 5.42          & 18.85          & 10.01 \\
ReSample  & 75.98          & 83.09          & 65.14          & 14.30          & 5.78          & 5.21           & 0.39 \\
DANN      & 74.11          & 81.83          & 66.11          & 15.73          & 5.56          & -1.60          & 3.50 \\
FairAdaBN & 74.70          & 82.02          & 65.13          & 14.10          & 5.95          & 1.33           & -1.67 \\
FairREAD  & 74.07          & 82.08          & \textbf{66.56} & 19.44          & 6.46          & -34.01         & -10.30 \\
Ours      & 75.47          & 83.36          & 65.34          & \textbf{10.86} & \textbf{5.01} & \textbf{28.26} & \textbf{14.53} \\
\noalign{\hrule height 0.8pt}
\end{tabular}
}
\end{table}
\vspace{-3mm}
\textbf{Representation-Level Fairness.}
Table~\ref{tab:attribute_prediction} evaluates demographic information retained in disease representations. We predict age, sex, and ethnicity from frozen disease embeddings by training multilayer perceptrons (MLPs). Our method achieves the lowest accuracy for the three attributes (69.43\%, 82.50\%, and 59.35\%) and the highest DRAR (59.04\%), indicating reduced demographic predictability and structural alignment with the demographic reference.
Figure~\ref{fig:feature_analysis} analyzes the learned representations for Pleural Effusion. Compared with ERM, our method shows greater overlap across demographic groups while preserving disease-label separation. The pairwise distances in the feature space show a consistent pattern, supporting improved cross-group alignment without compromising disease discrimination.
\begin{table}[!t]
\centering
\caption{Representation-level fairness results (\%). Demographic prediction accuracy is reported as mean $\pm$ standard deviation across four tasks. ``--'' denotes methods using demographic attributes at inference.}
\label{tab:attribute_prediction}
\fontsize{9}{11}\selectfont
\setlength{\tabcolsep}{2.5pt}
\renewcommand{\arraystretch}{1.15}
\begin{tabular*}{\columnwidth}{@{\extracolsep{\fill}}lcccc@{}}
\noalign{\hrule height 0.8pt}
& \multicolumn{3}{c}{Demographic Prediction Accuracy $\downarrow$} & \\
\cline{2-4}
Method & Age & Sex & Ethnicity & DRAR $\uparrow$ \\
\hline
ERM       & 72.32$\pm$1.66 & 86.22$\pm$2.12 & 62.00$\pm$1.34 & 0.00 \\
GroupDRO  & 71.98$\pm$1.73 & 85.63$\pm$2.01 & 61.51$\pm$1.48 & 22.22 \\
ReSample  & 72.20$\pm$1.90 & 85.80$\pm$2.44 & 61.99$\pm$1.40 & 2.81 \\
DANN      & 70.94$\pm$2.27 & 85.95$\pm$1.83 & 61.94$\pm$1.13 & 20.92 \\
FairAdaBN & -- & -- & -- & -0.60 \\
FairREAD  & -- & -- & -- & 13.07 \\
Ours      & \textbf{69.43}$\boldsymbol{\pm}$\textbf{4.09} & \textbf{82.50}$\boldsymbol{\pm}$\textbf{3.70} & \textbf{59.35}$\boldsymbol{\pm}$\textbf{1.61} & \textbf{59.04} \\
\noalign{\hrule height 0.8pt}
\end{tabular*}
\end{table}

\textbf{Ablation Study.}
Table~\ref{tab:ablation} evaluates the effects of dual-level decorrelation and prototype-guided cross-group alignment on disease and demographic representation learning. Removing prototype-guided alignment increases $\Delta_{\mathrm{EO}}$ from 10.86\% to 14.28\%, while removing decorrelation raises it to 17.72\% and reduces AUC to 82.36\%. Combining three components yields the best fairness and diagnostic performance.

\vspace{-5mm}
\begin{table}[h]
\centering
\caption{Ablation study of the proposed components.}
\label{tab:ablation}
\setlength{\tabcolsep}{2.5pt}
\renewcommand{\arraystretch}{1.05}
\resizebox{\columnwidth}{!}{
\begin{tabular}{lcccccccc}
\toprule
Method & \shortstack{$\mathcal{L}_{\mathrm{disease}}$\\$+\mathcal{L}_{\mathrm{attribute}}$} & \shortstack{$\mathcal{L}_{\mathrm{decorr}}$\\$+\mathcal{L}_{\mathrm{dCor}}$} & \shortstack{$\mathcal{L}_{d\text{-}con}$\\$+\mathcal{L}_{\mathrm{proto}}$} & BACC$\uparrow$ & AUC$\uparrow$ & F1$\uparrow$ & $\Delta_{\mathrm{EO}}\downarrow$ & $\Delta_{\mathrm{AUC}}\downarrow$ \\
\midrule
Ablation 1 & \checkmark & & & 73.59 & 82.21 & 62.60 & 16.67 & 5.30 \\
Ablation 2 & \checkmark & \checkmark & & 75.43 & 83.10 & 64.49 & 14.28 & 5.21 \\
Ablation 3 & \checkmark & & \checkmark & 75.15 & 82.36 & 64.09 & 17.72 & 5.79 \\
Full model & \checkmark & \checkmark & \checkmark & \textbf{75.47} & \textbf{83.36} & \textbf{65.34} & \textbf{10.86} & \textbf{5.01} \\
\bottomrule
\end{tabular}}
\vspace{-2mm}
\end{table}

\vspace{-3mm}
\section{Conclusion}
\label{sec:conclusion}
\vspace{-2mm}
We present a single-encoder framework for fair CXR classification across intersectional demographic groups. The framework disentangles disease and demographic representations through dual-level decorrelation and aligns disease representations using contrastive learning. Results across four CheXpert tasks demonstrate improved intersectional fairness at prediction and representation levels. These findings highlight the importance of reducing demographic dependence in learned representations for fair CXR diagnosis.

\newpage
\section{Acknowledgment}
This work is supported by the National Natural Science Foundation of China (NSFC) under Grant No. 82604003 and the XJTLU Research Development Fund (RDF-25-01-067). We acknowledge the CheXpert dataset used in this study. 

\bibliographystyle{IEEEbib}
\bibliography{strings,refs}

\end{document}